\documentclass{article} 
\usepackage{iclr2020_conference,times}
\usepackage{bbm}
\usepackage{graphicx}

\usepackage{amsmath,amsfonts,bm}

\def\eqref#1{equation~\ref{#1}}

\def\1{\bm{1}}

\DeclareMathAlphabet{\mathsfit}{\encodingdefault}{\sfdefault}{m}{sl}
\SetMathAlphabet{\mathsfit}{bold}{\encodingdefault}{\sfdefault}{bx}{n}

\usepackage{mathtools}

\usepackage{hyperref}
\usepackage{url}

\title{Uncertainty Estimation in Cancer Survival Prediction}

\author{Hrushikesh Loya\thanks{Corresponding Author: hrushikesh.loya@iitb.ac.in} \hspace{0.5mm}\textsuperscript{1}, Pranav Poduval\textsuperscript{1}, Deepak Anand\textsuperscript{1}, Neeraj Kumar\textsuperscript{2}, Amit Sethi\textsuperscript{1}\hspace{0.5mm}\textsuperscript{3}\\ 
\textsuperscript{1}Department of Electrical Engineering, Indian Institute of Technology Bombay \\
\textsuperscript{2}Department of Biomedical Engineering, Case Western Reserve University \\
\textsuperscript{3}Department of Pathology, University of Illinois at Chicago
}
\iclrfinalcopy 
\begin{document}

\maketitle

\begin{abstract}
Survival models are used in various fields, such as the development of cancer treatment protocols. Although many statistical and machine learning models have been proposed to achieve accurate survival predictions, little attention has been paid to obtain well-calibrated uncertainty estimates associated with each prediction. The currently popular models are opaque and untrustworthy in that they often express high confidence even on those test cases that are not similar to the training samples, and even when their predictions are wrong. We propose a Bayesian framework for survival models that not only gives more accurate survival predictions but also quantifies the survival uncertainty better. Our approach is a novel combination of variational inference for uncertainty estimation, neural multi-task logistic regression for estimating nonlinear and time-varying risk models, and an additional sparsity-inducing prior to work with high dimensional data.
\end{abstract}

\section{Introduction}

Survival modeling is key to precision oncology wherein cancer management, and treatment planning is personalized to a patient’s clinical, pathological, demographical, and genomic characteristics. Aided by the digitization of medical records, several studies over the past four decades have collected survival data based on longitudinal follow-up for various patient cohorts. Modeling survival in a cohort based on covariates known at the time of prediction is a complex task because the covariates to be taken into account can be large in number. These covariates can be entangled with each other by their interdependencies and interactions. A good survival model should give both (a) accurate survival estimates, and (b) a well-calibrated measure of uncertainty. We address the second problem in this work, which has escaped the attention it deserves.
 
Survival models have evolved primarily for accuracy over the last few decades. Cox proportional hazards model (Cox-PH) proposed by \cite{10.2307/2985181} is one of the oldest and most popular statistical models to predict survival. Cox-PH uses a hazard function to model the survival in a cohort and assumes that a patient’s relative log-risk of treatment failure (disease recurrence or death) at any time is a linear combination of the patient’s covariates that scales the underlying hazard function, which is another restrictive assumption. Multi-task logistic regression (MTLR) was proposed by \cite{NIPS2011_4210} as a remedy to the assumption of temporal constancy of relative risk between two patients, which led to increased prediction accuracy over Cox-PH. MTLR uses multi-task regression and joint likelihood minimization to model log-risk in a given time interval as a linear combination of the covariates. Recently, neural-MTLR was proposed by \cite{fotso2018deep} to move away from the linearity assumption as well to increase the prediction accuracy. Neural-MTLR models nonlinear interactions among covariates as features extracted by the lower layers of a neural network, whose last layers are the same as that of an MTLR model.

Most of the survival models aim for high accuracy on test cohorts that are very similar to their training cohorts. Consequently, even when their prediction is off on the test inputs or the test inputs are outliers with respect to the training input data distribution, they may predict (wrongly) with high confidence. For example, the above mentioned survival models are unable to access per-patient uncertainty in survival predictions. Uncertainty calibration is important if survival prediction models are to be deployed in clinical settings. The prediction of any model is usually untrustworthy when the test data from a new patient is out of the training distribution (OOD). In such OOD cases, it is important to involve human experts, and hence it is important to identify such cases with the model rightfully expressing high uncertainty or low confidence. Bayesian neural networks (BNNs) provide a framework to capture the underlying uncertainties inherent to both the data (data uncertainty) and the limitations of the model (model uncertainty). We propose a Bayesian extension of (neural) MTLR that can capture patient-specific survival uncertainties. Capturing uncertainty in the prediction also helps us handle heterogeneous data and analyze prognostically important covariates. Furthermore, we incorporate a prior in our model to sparsify a large number of input covariates.

\section{Theoretical Background and Proposed Method}


\subsection{Survival models}

The setting that we assume has a set of covariates $x_i$ associated with each patient $i$, a time to adverse event ($T_i$) (usually death or disease recurrence), and an event indicator ($E_i$). The event indicator $E_i=1$ means that the patient died after a time interval of $T_i$. Patients with $E_i=0$ are called right-censored, indicating that she was surviving (or living disease-free) at time $T_i$, but survival beyond that time in unknown.

The survival function and hazard function are two important outcomes of survival models. The survival function, $S(t) = P(T \geq t)$, is the probability of a patient to survive more than time t. The hazard function $\lambda(t)$ is given by $\lim_{\Delta t \to 0} P(t \geq T \geq t+\Delta t \mid T \geq t)/\Delta t$, which means the probability that an individual will not survive an extra infinitesimal amount of time $\Delta t$, given they have already survived up to time t. Cox-PH \citep{10.2307/2985181} models the hazard function $\lambda(t|\vec{x})$ at time $t$ for a given vector input covariates $\vec{x}$ in terms of an underlying hazard function $\lambda_{0}(t)$ and linear weights $\theta$ for the covariates as follows: $\lambda(t|\vec{x}) = \lambda_{0}(t) \exp (\vec{x}^{T}\vec{\theta})$.

MTLR \citep{NIPS2011_4210} assumes a series of logistic regression models for $m+1$ time intervals, where $m$ is chosen based on the desired fineness of temporal variation and the size of the training data, as follows: $P_{\theta_{i}}(T \geq t_i \mid x) = (1 + exp(\vec{\theta}_i . \vec{x} + b_i))^{-1}; 0 \leq i \leq m$. The parameters $\vec{\theta_i}$ and $b_i$ depend on the time interval $i$, whereas the input vector $\vec{x}$ is same for all regression models. However, the outputs of these logistic regression models are not independent, because a death event at time $t_i$ would mean a death event at all subsequent time points $t_j, j > i$. We encode the output of the regression model, using a $m$-dimensional binary sequence $y = (y_1, y_2, y_3 ... y_m)$, where, $y_i = 0$ means that the patient is living at time $t_i$ and $y_i = 1$ means that the patient is dead at time $t_i$. Thus, once we encounter a $y_i = 1$, all subsequent $y_j, j > i$ are bound to be 1. A smoothness prior on the parameters across time ensures that the predictions are not noisy. The probability of observing a sequence $y = (y_1, y_2, y_3 ... y_m)$ is the likelihood of the model. It can be generalized by the logistic regression model as follows: $P_{\theta}( Y = (y_1, y_2, y_3 ... y_m) \mid \vec{x}) = \exp \left[\sum_{k=j}^{m}y_i (\vec{\theta}_i . \vec{x} + b_i)\right] / \left[\sum_{k=0}^{m} \exp (f_{\theta} (\vec{x},k))\right]$, where $f_{\theta} (\vec{x},k) = \sum_{i=k+1}^{m} (\vec{\theta}_i . \vec{x} + b_i)$.

The MTLR loss function for uncensored patients is obtained by taking the logarithm of the joint likelihood term and adding regularization terms for temporal smoothness of the parameters and the resultant predictions, as follows:
\begin{equation}
\label{eq:MTLR_loss}
    L = \frac{C_1}{2} \sum_{j=1}^{m} {\Vert \vec{\theta}_j \Vert }^2 + \frac{C_2}{2} \sum_{j=1}^{m} {\Vert \vec{\theta}_{j+1} - \vec{\theta}_j \Vert }^2 - \sum_{i=1}^{n} \left[ \sum_{j=1}^{m} y_j(s_i) ({\theta}_j . \vec{x} + b_j) - \log( \sum_{k=0}^{m} exp (f_{\theta} (\vec{x}_i,k)) \right]
\end{equation}
where $C_1$ and $C_2$ are hyperparameters that control the amount of smoothing in the parameters and $n$ is the number of patients.

For right-censored patients (those who are lost to follow-up), there are more than one consistent binary sequences of $y_i$'s. In this case the likelihood of the patient is the sum of likelihoods of all possible sequences. The overall likelihood for censored patients whose last contact was closest to time point $t_j$ is given as follows: $P_{\theta_{i}}(T \geq t_i \mid x) = \left[\sum_{k=j}^{m}exp (f_{\theta} (\vec{x},k))\right] / \left[\sum_{k=0}^{m} exp (f_{\theta} (\vec{x},k))\right]$.

Neural-MTLR \citep{fotso2018deep} models nonlinear combinations of the covariates as inputs to the MTLR model, where both the MTLR model and the nonlinear feature extraction are trained end-to-end using backpropagation (gradient descent) on a loss function similar to Equation~\ref{eq:MTLR_loss}.

\subsection{Variational Inference}

A feed-forward neural network trained with gradient descent will arrive at point estimates. However, in the case of Bayesian NNs (BNNs) \citep{neal2012bayesian}, the weights are not point estimates but a parameterized probability distribution. Our task is to find a distribution over the parameters given the input data, i.e., $p(\theta \mid D)$. With this posterior, we can predict test output $y^{*}$ for a new test input $x^{*}$ by marginalizing the likelihood over the parameters $\theta$. However, even for the modest-sized NNs, the number of parameters prohibits an analytical calculation of uncertainty, and one has to resort to approximate inference methods. We define an approximating variational distribution $q_{\psi}(\theta)$ with parameters $\psi$. Then the Kullback-Leibler divergence (KL) with respect to the parameters $\psi$ is minimized between the proposed posterior and the true posterior.

Minimizing the KL divergence is equivalent to minimizing the variational free energy \citep{friston2007variational}, \citep{blundell2015weight}, where the latter is often computed on $M$ mini batches $D^1, D^2, \dots, D^M$ for computational tractability. We then estimate the cost using an unbiased Monte Carlo (MC) approximation for each mini batch as follows: 

$\theta^j \sim q_{\psi}(\theta), \quad \Hat{\mathcal{L}}(\psi) = - \sum\limits_{i\in D^j} \log(p(y_{i}|f^{\theta^j}(x_{i}))) + (1/M) \text{KL}(q_{\psi}(\theta)||p(\theta))$.

\subsection{Proposed probabilistic weights to model uncertainty}

We assume the posterior and the prior on weights to be a spike and slab, which is standard for sparse linear models \citep{doi:10.1080/01621459.1988.10478694} \citep{doi:10.1080/01621459.1993.10476353} \citep{NIPS2011_4305}. Recently, a closed form expression for the KL divergence between the spike and slab posterior and spike and slab prior was derived by ~\cite{tonolinivariational}, which we utilized in this work. The prior probability density is given as follows: $ p(\theta) =  \prod_{i=1}^{N} (\alpha \mathcal{N}(\theta_i;0,1) + (1-\alpha) \delta(\theta_i))$, where $\delta(.)$ is the dirac delta function centered at zero. The sparsity of solution can be increased for this prior by decreasing $\alpha$ from one to zero. The posterior is chosen to be of similar form, given as: $q_{\psi}(\theta) = \prod_{i=1}^{N} (\gamma_i \mathcal{N}(\theta_i;\mu_i,\sigma_i^2) + (1 - \gamma_i) \delta(\theta_i))$, where $\mu_i$, $\sigma_i$ and $\gamma_i$ are the free parameters of the neural network. The choice of posterior not only allows us to derive an analytical lower bound for the KL divergence between assumed posterior and prior but also gives additional degree of freedom compared to a fully factorized Gaussian. 

In order to quantify data uncertainty, we use the standard trick of predicting not only mean but also the variance of survival probability \citep{kendall2017uncertainties}. Our overall prediction now becomes a sample drawn from this Gaussian, as follows: $ y_{out} = \hat{y} + \hat{\sigma}.\epsilon; \epsilon\sim N(0,1)$, where $\hat{y}$ and $\hat{\sigma}$ are approximated using Monte Carlo samples. The loss function for the mini batch $D^i$ of our Bayesian variant is given as follows:
\begin{equation}
 \Hat{\mathcal{L}}(\psi) = -\log p(D^i|\theta^{i}) +\frac{1}{M} \sum_{i=1}^{N} \left(  \frac{\gamma_i}{2}(\mu_i^2 + \sigma_i^2 - \log( \sigma_i^2)) \\ 
 + (1-\gamma_i)\log(\frac{1 - \alpha}{1 - \gamma_i}) + \gamma_i\log(\frac{\alpha}{\gamma_i}) \right ) 
\end{equation}
\begin{equation*}
\theta^i \sim q_{\psi}(\theta),  \end{equation*}
where $-\log p(D^i|\theta^{i})$ is the negative log-likelihood defined as in equation \ref{eq:MTLR_loss}, $M$ is the number of mini-batches and $N$ is the total number of parameters. One can see that setting $\alpha = 1$ and $\gamma = 1$ reduces this expression to a fully factorized Gaussian posterior and prior that is used in varational autoencoders \citep{kingma2013autoencoding}.

\begin{figure}
    \centering
    \includegraphics[width = 90mm]{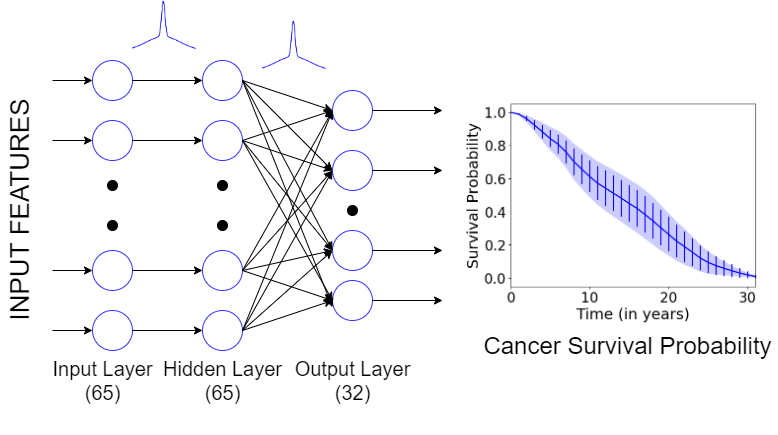}
    \caption{Proposed neural network architecture with weights sampled from the spike and slab posterior give survival probability in solid curve, along with data uncertainty ($\pm$ 1 std. dev.) in vertical bars, and model uncertainty ($\pm$ 1 std. dev.) in shaded region.}
    \label{fig:BBB-NMTLR-arch}
\end{figure}

We used a simple one-hidden layer Bayesian neural network with spike and slab prior and posterior, and ReLU activation in all but the final layer. The number of inputs to the network are equal to the number of covariates, and the number of outputs equal to number of time intervals. Instead of a fully connected structure from the first layer to the hidden layer, we only assume a one-to-one mapping to simulate variable elimination based on the sparsity inducing prior, as shown in Figure \ref{fig:BBB-NMTLR-arch}.   

\section{Results}
 
Using a subset of 47 out of the PAM50 gene expressions and clinical variables that were common to both TCGA-BRCA \citep{TCGA} and METABRIC \citep{curtis2012genomic} datasets we trained on one dataset and tested on the other to obtain results on model accuracy. We combined both datasets and held out samples at random for experiments on variable importance and uncertainty estimation.

\subsection{Survival predictions}

C-index and Integrated Brier Score (IBS) are two commonly used metrics for analyzing the accuracy of survival models for censored data, where the former is a generalization of the area under the ROC curve (AUC), and the latter is the average weighted squared distance between the observed and predicted survival. Thus, a higher C-index and lower IBS implies a more accurate model. Table \ref{tab 1} shows our method performs better compared to Cox-PH, MTLR, and a comparable neural-MTLR model with a single hidden layer.

\begin{table}[h]
    \centering
\caption{Comparison of mean ($\pm$ std. dev.) C-index and IBS across survival models using one of TCGA-BRCA and METABRIC datasets for training and the other for testing.}
\vspace{2mm}
\begin{tabular}{|c|c|c|}
\hline
    Methods & C-index &  IBS\\ \hline
    Cox-PH & 0.65 $\pm$ 0.10 & 0.20 $\pm$ 0.07 \\\hline
    MTLR & 0.68 $\pm$ 0.06 & 0.21 $\pm$ 0.06 \\\hline
    N-MTLR & 0.68 $\pm$ 0.02 & 0.16 $\pm$ 0.04\\\hline
    Our Method & \textbf{0.71 $\pm$ 0.05} & \textbf{0.12 $\pm$ 0.02} \\\hline
\end{tabular}
\label{tab 1}
\end{table}

\subsection{Ranking prognostic features}

\begin{figure}[h]
    \centering
    \includegraphics[width=120mm, trim={3.0cm 18.6cm 2.8cm 2.8cm},clip]{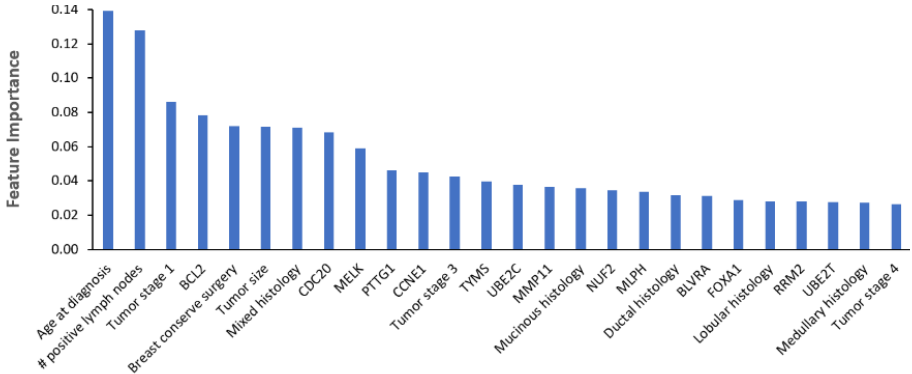}
    \caption{Importance scores for a truncated list of the features}
    \label{fig:feature_importance}
\end{figure}

We obtained feature importance for each input feature based on the distribution of weights learned by the network from the first layer to the hidden layer. We interpreted the ratio of mean and standard deviation of the weight associated with a feature as its signal to noise ratio. In case of spike and slab posterior, the signal to noise ratio for feature i is given by: $\mid \mu_i \mid / (\sigma_i * \gamma_i)$. We observe in Figure~\ref{fig:feature_importance} that age at diagnosis, lymph node metastasis, and tumor stage are among the top three prognostically important features. Among the genomic signatures, BCL2 is an antiapoptotic protein whose prognostic role in breast cancer is consider as  sub-type specific. It is a good prognostic marker mostly for Luminal A breast cancers~\citep{corces2018chromatin}. CDC20 is an oncoprotien that promotes the development and progression of breast cancer, and its overexpression is often associated with poor short-term survival, specifically in triple-negative breast cancers~\citep{karra2014cdc20}. CDC25L (a.k.a RASGRF1) plays a key role in tumor cell proliferation and and inflammation through mitogen-activated protein kinase (MAPK) pathway in breast cancers~\citep{rodrigues2012angiotensin}. Similarly, PTTG1 gene enhances the migratory and invasive properties of breast cancer cells by inducing epithelial to mesenchymal transition~\citep{tetreault2013kruppel}. Some of these genes have not found a regular place in risk assessment assays, whereas our method may be able to systematically suggest their direct association with survival.


\subsection{Low confidence on out-of-distribution (OOD) test data}

\begin{figure}[h]
    \centering
    \includegraphics[width = 50mm]{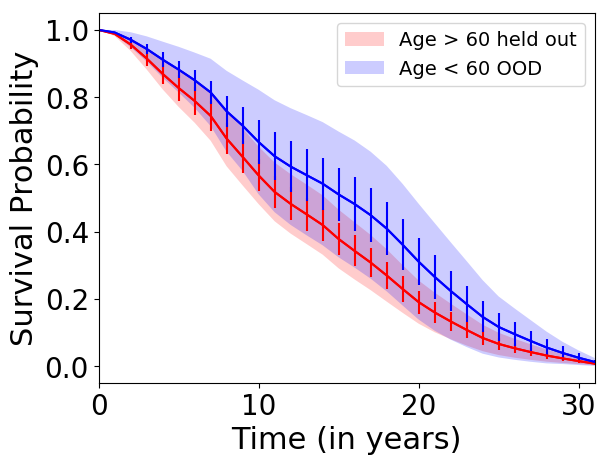}
    \includegraphics[width = 50mm]{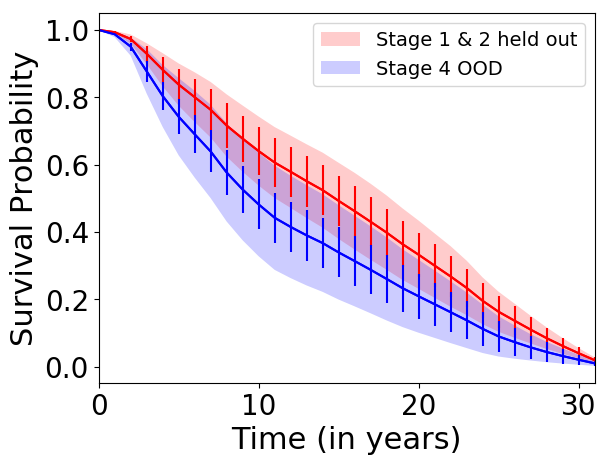}
    \caption{Predicted survival (curve) and model uncertainty (shaded area) for held-out and OOD data.}
    \label{fig:uncertainty}
\end{figure}

In order to demonstrate the use of quantifying uncertainty, we divided the entire data (TCGA + METABRIC) into old (age $>$ 60 years) and young patients (age $<$ 60 years), where 60 years is the median age of patients in the dataset. We trained the model on 80\% of the old patients and tested it on the remaining 20\% old as well as all of the young patients. We define mean uncertainty score associated with a survival prediction as the mean of the standard deviations in model predictions (for 50 forward passes) across all time points. The test cohort of younger patients was successfully identified as OOD as their mean uncertainty was 110\% higher than that of the test cohort of older patients, as shown in Figure~\ref{fig:uncertainty}). Similarly, we trained another model on a subset of patients with low cancer stage and saw a 43\% higher mean uncertainty score for higher-stage patients (OOD) as compared to the held-out lower stage patients.     

\section{Conclusion}

We proposed a Bayesian framework for modeling survival prediction that not only gives more accurate predictions but is also able to select prognostically important features in the data and detect test samples that are out of the training distribution. This makes our model more interpretable and trustworthy due to its well-calibrated uncertainty estimates. Our approach is a step in the direction of training models that go beyond a singular focus on test prediction accuracy to that of recognizing uncertainty appropriately in new cohorts and producing new biological insights. Such approaches should be further tested in larger multi-institutional and multi-cohort settings. 



\newpage

\bibliography{iclr2020_conference}
\bibliographystyle{iclr2020_conference}

\end{document}